\documentclass{article}
\usepackage{arxiv}
\usepackage[utf8]{inputenc} 
\usepackage[T1]{fontenc}    
\usepackage[pagebackref,breaklinks,colorlinks]{hyperref}
\usepackage{url}            
\usepackage{booktabs}       
\usepackage{amsfonts}       
\usepackage{nicefrac}       
\usepackage{microtype}      
\usepackage{lipsum}
\usepackage{graphicx}
\graphicspath{ {../LaTeX/} }
\usepackage{graphicx}
\usepackage{amsmath}
\usepackage{amssymb}
\usepackage{booktabs}
\usepackage{wrapfig}
\usepackage{comment}
\usepackage{multirow}
\usepackage[ruled, vlined, linesnumbered]{algorithm2e}%
\usepackage{algpseudocode}
\usepackage{amsmath}
\usepackage{amsthm}
\usepackage{bbm}
\usepackage[ruled,vlined]{algorithm2e}
\usepackage{mathtools}

\newcommand{\gcrpp}{\textsc{GCR}}
\newcommand{\derpp}{\textsc{DER}}
\newcommand \doubleplus{\mathbin{+\mkern-10mu+}}

\usepackage[capitalize]{cleveref}
\crefname{section}{Sec.}{Secs.}
\Crefname{section}{Section}{Sections}
\Crefname{table}{Table}{Tables}
\crefname{table}{Tab.}{Tabs.}

\title{GCR: Gradient Coreset based Replay Buffer Selection for Continual Learning}

\author{
  Rishabh Tiwari~\textsuperscript{1,3}, Krishnateja Killamsetty~\textsuperscript{2}, Rishabh Iyer~\textsuperscript{2}, Pradeep Shenoy~\textsuperscript{3} \\
  {\small \textsuperscript{1}Department of Physics,  Indian Institute of Technology (ISM) - Dhanbad} \\
 {\small \textsuperscript{2}Department of Computer Science, University of Texas at Dallas}\\
  {\small \textsuperscript{3}Google Research, India}\\
  \texttt{ \small \{rishabhtiwari,shenoypradeep\}@google.com, \{krishnateja.killamsetty,rishabh.iyer\}@utdallas.edu}
}
\author{
  Rishabh Tiwari \\
  Google Research, India \\
  \texttt{rishabhtiwari@google.com} \\
  \And
  Krishnateja Killamsetty \\
  Department of Computer Science\\
  University of Texas at Dallas\\
  \texttt{krishnateja.killamsetty@utdallas.edu}\\
  \And
  Rishabh Iyer \\
  Department of Computer Science\\
  University of Texas at Dallas\\
  \texttt{rishabh.iyer@utdallas.edu} \\
  \And
  Pradeep Shenoy \\
  Google Research India\\
  \texttt{shenoypradeep@google.com} \\
}

\begin{document}
\maketitle
\begin{abstract}
Continual learning (CL) aims to develop techniques by which a single model adapts to an increasing number of tasks encountered sequentially, thereby potentially leveraging learnings across tasks in a resource-efficient manner. A major challenge for CL systems is catastrophic forgetting, where earlier tasks are forgotten while learning a new task. To address this, replay-based CL approaches maintain and repeatedly retrain on a small buffer of data selected across encountered tasks. We propose Gradient Coreset Replay (\gcrpp), a novel strategy for replay buffer selection and update using a carefully designed optimization criterion. Specifically, we select and maintain a 'coreset' that closely approximates the gradient of all the data seen so far with respect to current model parameters, and discuss key strategies needed for its effective application to the continual learning setting. We show significant gains (2\%-4\% absolute) over the state-of-the-art in the well-studied offline continual learning setting. Our findings also effectively transfer to online / streaming CL settings, showing up to 5\% gains over existing approaches.  Finally, we demonstrate the value of supervised contrastive loss for continual learning, which yields a cumulative gain of up to 5\% accuracy when combined with our subset selection strategy.
\end{abstract}


\section{Introduction}

The field of continual learning (CL)~\cite{thrun1995lifelong} studies the training of models in an incremental fashion, to  generalize across a number of sequentially encountered scenarios or tasks, and to avoid the training \& maintenance costs of one-off models. A key challenge in CL is the limited access to data from prior tasks; this results in \textit{catastrophic forgetting}~\cite{mccloskey1989catastrophic}, where training on subsequent tasks may potentially erase information in the model parameters pertaining to previous tasks. Approaches to address catastrophic forgetting\footnote{The papers cited, here and subsequently, under each paradigm or technique are representative or recent papers, not necessarily the papers defining or proposing said paradigm.} include modifications to the loss function (e.g.,~\cite{kirkpatrick2017overcoming,Rudner2021cfsvi}), to the network architecture (e.g.,~\cite{hung2020compacting,SOKAR2021}), and to the training procedure and data augmentation (e.g.,~\cite{derpp}).  In particular, \textit{replay-based} continual learning maintains a small data sketch from previous tasks, to be included in the training mix throughout the lifetime learning of the model. Remarkably, as little as 1\% of saved historical data, using random sampling, is enough to provide significant gains over other CL approaches~\cite{derpp}. This suggests that  sophisticated methods for selecting compact data summaries or \textit{coresets} may perform much better. However, previous approaches to coreset in continual learning have focused on qualitative/diversity-based criteria~\cite{aljundi2019gradient,ocs}, or bi-level optimizations with significant computational costs and scaling limitations~\cite{borsos2020coresets}.

\begin{wrapfigure}{r}{0.5\textwidth}
    \centering
    \includegraphics[width=0.4\textwidth, height=4cm]{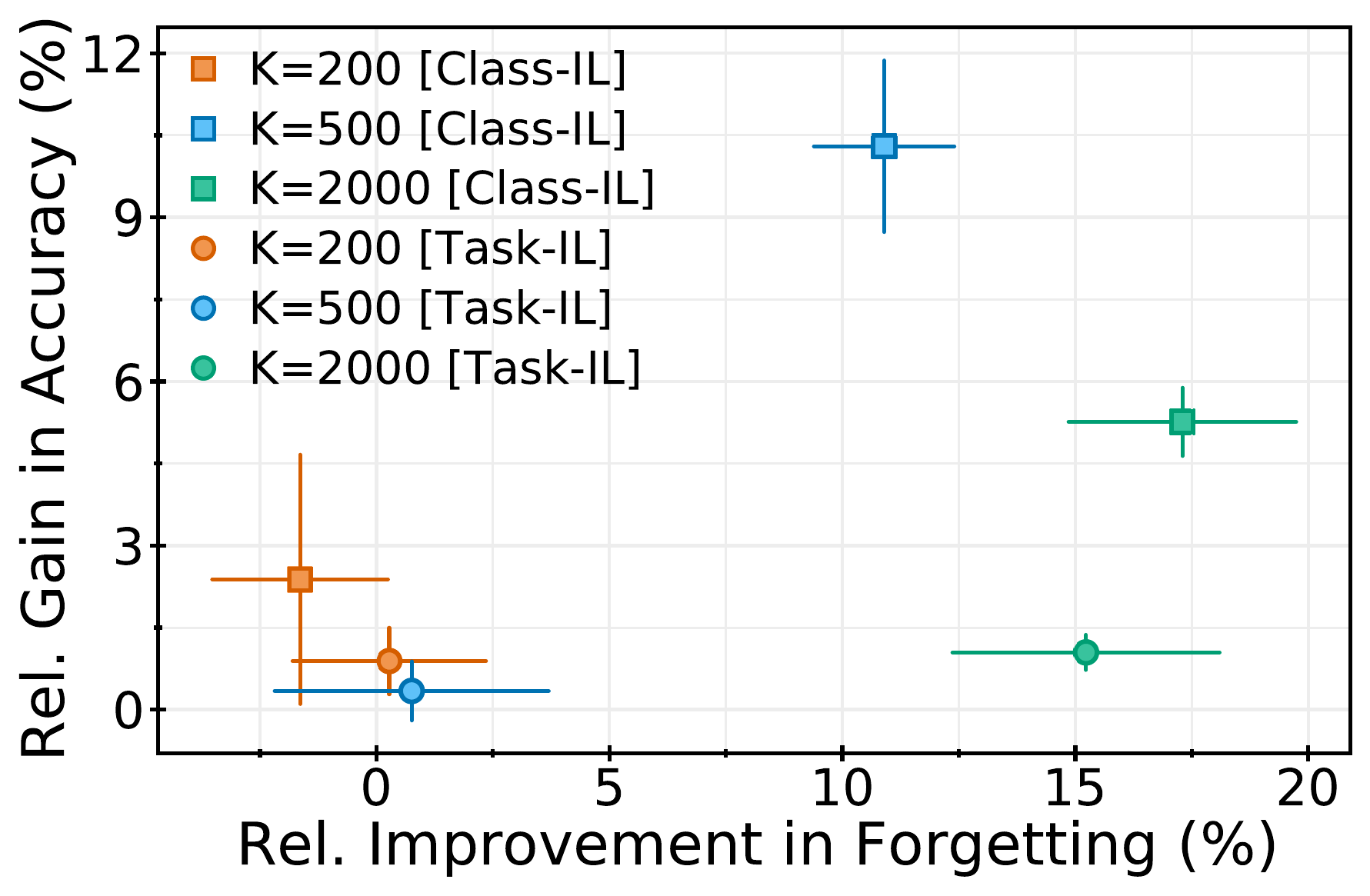}
    \caption{When training CL models on S-Cifar100 with different replay buffer sizes $K \in [200, 500, 2000]$, the use of replay buffers selected by \gcrpp\ produces higher model accuracy and lower model forgetting than using reservoir-sampled replay buffers~\cite{derpp}. See text for more details.}
    \label{fig:main_perf}
\end{wrapfigure}

We present Gradient-based Coresets for Replay-based CL (\gcrpp), a principled, optimization-driven criterion for selecting and updating coresets for continual learning. Specifically, we select a coreset that approximates the gradient of model parameters over the entirety of the data seen so far. We provide empirical evidence that coresets selected utilizing the above approach mitigate catastrophic forgetting on par with, or better than previous methods. Figure~\ref{fig:main_perf} illustrates how our coreset selection approach improves over random data selection by large margins, both in overall accuracy and in  retaining of previous tasks. We explore the effect of better representation learning in the CL setting by including a representation learning component~\cite{supcon} into our CL loss function. We conduct extensive experimentation against the state-of-the-art (SOTA) in the well-studied offline CL setting, where the current task’s data is available in entirety for iterative training (2-4\% absolute gains over SOTA), as well as the online/streaming setting, where the data is only available in small batches and cannot be revisited (up to 5\% gains over SOTA). In particular, we show that the benefit from our coreset selection mechanism increases with increasing number of tasks, showing that GCR scales effectively with task count. Finally, we demonstrate the superiority of \gcrpp\ over other coreset methods for CL in a head-to-head comparison.

\section{Related Work}

CL approaches divide broadly into three main themes:\\
\noindent {\bf Regularization based approaches} preserve learning from earlier tasks via regularization terms placed on model weights (EWC~\cite{Kirkpatrick3521}, synaptic intelligence~\cite{synaptic}), structural regularization~\cite{pmlr-v80-serra18a}, or functional regularization~\cite{Cha2021,Rudner2021cfsvi}). In other work, incremental momentum matching~\cite{NIPS2017_f708f064} proposes a model-merging step which merges a model trained on the new task with the model trained over the previous tasks, or use knowledge distillation-based regularization~\cite{L2F}  to retain learnings from previous tasks.

\noindent \textbf{Architecture adjustments} modify model architectures to incorporate CL challenges; for instance, using a recurrent neural network~\cite{9207550},  learning overcapacitated deep neural networks in an adaptive way that repeatedly compresses sparse connections~\cite{SOKAR2021} or using overlapping convolutional filters~\cite{hung2020compacting}. Other work attempts to identify shared information across tasks (CLAW~\cite{Adel2020Continual}) using variational inference.

\noindent {\bf Replay based approaches} preserve knowledge on previous tasks by storing a small memory buffer of representative data samples from previous tasks for continued training alongside new tasks (ER~\cite{doi:10.1080/09540099550039318, Ratcliff90connectionistmodels}). Some approaches add a distillation loss to preserve learned representations (iCaRL~\cite{rebuffi2017icarl}) or for classifier outputs (\derpp~\cite{derpp} for points in the buffer. Gradient Episodic Memory~(\textbf{GEM})~\cite{lopez2017gradient}, \textbf{AGEM})~\cite{chaudhry2019efficient}) focuses on minimizing  catastrophic forgetting by the efficient use of episodic memory. Meta-Experience Replay~(\textbf{MER})~\cite{riemer2018learning} adds a penalty term for between-task interference in a meta-learning framework. Maximally Interfered Retrieval~(\textbf{MIR}) method~\cite{MIR} uses controlled memory sampling for replay buffer by retrieving samples that the foreseen model parameters update will most negatively impact. Of these approaches, the distillation loss on the replay buffer~\cite{rebuffi2017icarl,derpp} show the strongest performance gains at standard CL benchmarks.

\noindent {\bf Coreset selection:} Coresets \cite{feldman2020core} are small, informative weighted data subsets that approximate specific attributes (e.g., loss, gradients, logits) of original data. Previous work~\cite{har2004coresets, Lemke2016DensitybasedCA} used coresets for unsupervised learning problems such as K-means \& K-medians clustering. Coresets enable efficient and scalable Bayesian inference~\cite{pmlr-v130-zhang21g, NEURIPS2019_7bec7e63, 10.5555/3322706.3322721, DBLP:conf/icml/CampbellB18, NIPS2016_2b0f658c} by approximating the model logit sum of the entire data. Several recent works~\cite{killamsetty21a, NEURIPS2020_8493eeac, Mirzasoleiman2020CoresetsFD, killamsetty2021glister, killamsetty2021retrieve}  also use coresets to approximate the gradient sum of the individual samples across the entire dataset, for efficient and robust supervised learning. Although coresets are increasingly applied to supervised and unsupervised learning scenarios, the problem of coreset selection for CL is relatively under-studied. We discuss existing coreset based approaches for CL below.\\
\noindent {\bf Coresets for Replay-based CL:} Approaches include selecting replay buffers by maximizing criteria like sample gradient diversity (\textbf{GSS}~\cite{aljundi2019gradient}), a mix of mini-batch gradient similarity and cross-batch diversity (\textbf{OCS}~\cite{ocs}), adversarial Shapley value(\textbf{ASER}~\cite{shim2021online}), or representation matching (iCARL~\cite{rebuffi2017icarl}).  These approaches introduce a secondary selection criterion (to reduce the interference of the current task and forgetting of previous tasks) that is highly specific to each particular approach. As a result, they may not generalize different CL loss criteria.
Additionally, previous approaches have not considered weighted replay buffer selection, which increases the representational capabilities of replay buffers. 
Closest to our work is a recent paper~\cite{borsos2020coresets} that proposed solving a bi-level optimization for selecting the optimal replay buffer under certain assumptions, rather than for any specified loss function. Their proposal is computationally expensive, intractable in large task scenarios, and scales poorly with buffer size; moreover, they did not compare against recent strong proposals~\cite{derpp}. We compare directly against this approach in our experiments. In contrast to previous coreset approaches in CL, (a) We use an optimization criterion directly tied to the replay loss function used by the CL approach; we show that doing so can achieve better model accuracies across different replay loss functions (ER~\cite{riemer2018learning} and DER~\cite{derpp}, see Results), (b) We use a weighted coreset selection mechanism for continual learning, where the weights  are selected by the coreset optimization criterion and allow effective use of the buffer data.\\
\noindent \textbf{CL settings:} Apart from different learning approaches, CL is also studied under a range of settings\footnote{see van de Ven \& Tolias~\cite{Ven2019ThreeSF} for a taxonomy}, depending on what information and data are available at which stage. In particular, the task-incremental and class-incremental settings assume that task label is known, or not available, respectively, at inference time. In the offline setting, each new task's data is available in entirety for repeated/ iterative learning~\cite{DeLangeMatthias2021Acls}, whereas online CL only considers data availability in small buffers that cannot be revisited~\cite{10.5555/304710.304720}. 

\begin{wrapfigure}{r}{0.45\textwidth}
  \includegraphics[width=0.4\textwidth,height=4cm]{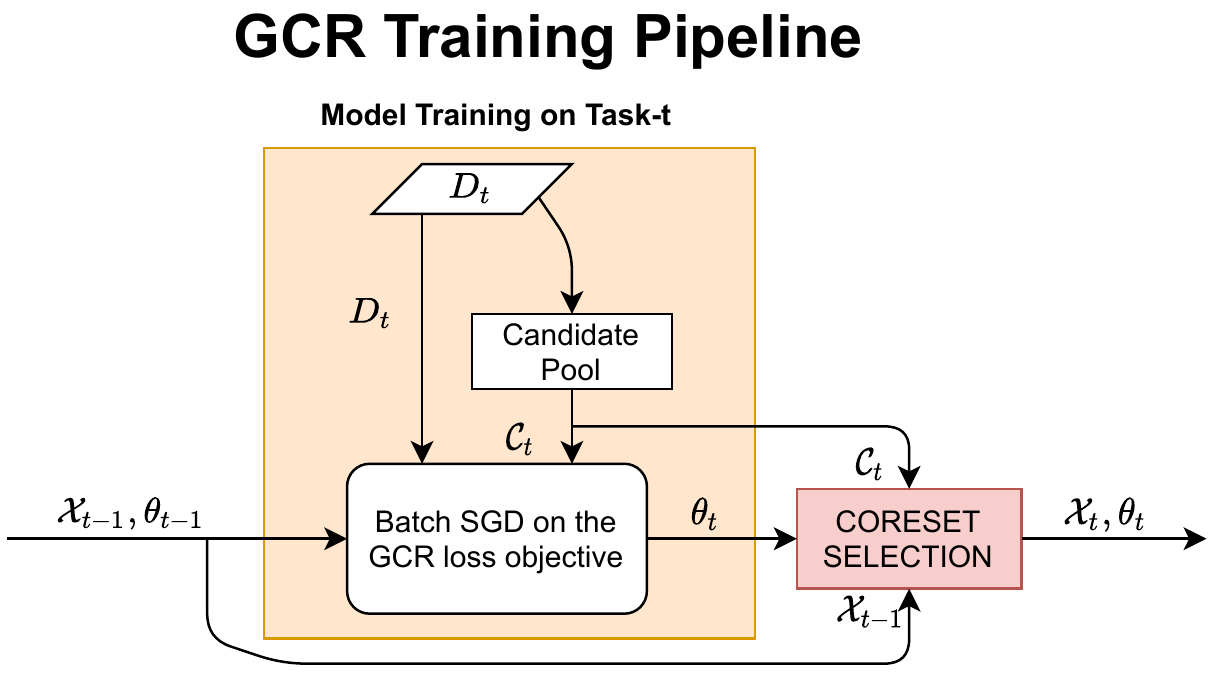}
  \caption{Block diagram showing training, replay buffer selection, and update operations of \gcrpp\ at time step $t$.}
  \label{fig:gcr_pipeline}  
\end{wrapfigure}

\section{Preliminaries}
\subsection{Notation}
We assume that there are $T$ tasks in the continual learning classification problem considered. For each task $t \in \{1, 2, \cdots , T\}$, we have an associated dataset $D_t$ that is composed of i.i.d data points  $\{(x_{it}, y_{it})_{i=1}^{|D_t|}\}$ where $x_{it}$ is the $i^{th}$ sample and $y_{it}$ is the ground truth label for the $i^{th}$ sample in the dataset $D_t$. We assume that each task $t$ is associated with a set of distinct classes $y_t = {y_{t1}, y_{t2}, \cdots, y_{tn}}$ and that no tasks have common classes i.e $\underset{t \ne k}{\forall} y_t \cap y_k = \emptyset$. Let the classifier model be characterized by parameters $\theta$. We split the model parameters into feature extraction layer and linear classification layer. Let the model's feature extractor output for input sample $x$ is denoted by $\Omega_{\theta}(x)$. The model logits output for input sample $x$ is denoted by $h_{\theta}(x)$ and the predicted probability distribution over the classes for input $x$ is denoted by $f_{\theta}(x) = \textsc{softmax}(h_{\theta}(x))$. We use $t$ to denote the last task observed so far. Let, $l$ be the classification loss function (such as cross-entropy loss).Let the data buffer of previous tasks used for the replay be denoted by $\mathcal{X}$ and $\mathcal{L}_{rep}(\theta, \mathcal{X})$ be the replay-buffer loss over the entire replay buffer $\mathcal{X}$.

\subsection{Continual Learning}
Following the above notations, the goal of continual learning at step $t$ is to minimize the following objective:
\begin{equation}
\small
\begin{aligned}
      \underset{\theta}{\operatorname{argmin\hspace{0.7mm}}} \sum_{i=1}^{t} \sum_{(x, y) \in D_{i}} l(y, f_{\theta}(x))
\end{aligned}
\end{equation}

where $l$ is the cross-entropy loss function.

In a continual learning setup, at step $t$, we only have access to data points from task $t$, making it difficult to optimize the above objective directly. In particular, we have to make sure that while the model is learning on a new task, it should not forget previous tasks.

\subsection{Replay-based Continual Learning}
Replay-based CL methods maintain a small buffer of data points from previous tasks on which the model is trained and the data samples from the new task to make the model retain the knowledge of previous tasks. Following the notation earlier, we denote the replay buffer of previous tasks by $\mathcal{X}$. One of the possible formulations for replay-based methods for continual learning is as follows:

\begin{equation}
\small
\begin{aligned}
    \underset{\theta}{\operatorname{argmin\hspace{0.7mm}}}  \sum_{(x, y) \in D_{t}} l(y, f_{\theta}(x)) + \lambda \mathcal{L}_{rep}(\theta, \mathcal{X})  
\end{aligned}
\label{eq:UNKNOWN}
\end{equation}

where $\mathcal{L}_{rep}$ is the replay loss of the model on samples from replay buffer, and $\lambda$ is a hyperparameter denoting the replay loss coefficient.  Early work used cross-entropy for the replay loss (see e.g., ER~\cite{doi:10.1080/09540099550039318}); however, recent work (Dark Experience Replay, \derpp~\cite{derpp}), proposes storing the logits ($z$) associated with data points at the time of selection for use in an additional distillation loss. The authors show that \derpp\ outperforms a range of previous proposals in both task-incremental and class-incremental offline continual learning scenarios. The  \derpp\ objective is:
\begin{equation}
\label{eq:derloss}
\small
\begin{aligned}
    &\mathcal{L}_{rep}(\theta, \mathcal{X}) = \sum_{(x, y) \in \mathcal{X}} \bigg(\alpha {\left \Vert z - f_{\theta}(x) \right \Vert }^2  + \beta l(y, f_{\theta}(x)) \bigg)
\end{aligned}
\end{equation}

\section{\gcrpp: Methods}
Figure~\ref{fig:gcr_pipeline} shows the overall \gcrpp\ workflow. We frame an incremental update process that takes the previously selected replay buffer $\mathcal{X}_{t-1}$ along with a \textit{candidate pool} $\mathcal{C}_t$ from the current task's data $D_t$. We work with the candidate pool $C_t$ instead of the task data $D_t$ in order to have a more general formulation that covers both Offline CL (all current task data is available) and Online CL (data arrives sequentially in small buffers) scenarios. 

Our primary contribution is a formulation of the \textit{replay buffer selection} as a principled optimization problem based on gradient approximation. We discuss the various components of \gcrpp\ needed to support this optimization objective, and then sketch how they combine in our overall continual learning framework.

\subsection{GradApprox for Replay Buffer Selection}
\label{sec:buffer_selection}

We propose a weighted gradient approximation objective for selecting replay buffers. Given a dataset $D=\{d_i\}_{i=1}^{i=|D|}$, associated sample weights $W_D=\{w_i\}_{i=1}^{i=|D|}$, and a subset size $K$, GradApprox selects a data subset $\mathcal{X} : \mathcal{X} \subset D, |\mathcal{X}| = K$ and its associated weights $W_{\mathcal{X}}$, using the following optimization objective:
\begin{equation}
\label{subsetobj}
\small
    \begin{aligned}
          &\underset{\mathcal{X}, W_{\mathcal{X}}: \mathcal{X} \subset D, |\mathcal{X}| = K, W_{\mathcal{X}} \geq 0}{\operatorname{argmin}} L_{sub}(D, W_D, \mathcal{X}, W_{\mathcal{X}}, \lambda) \text{  where } \\
          &L_{sub}(D, W_D, \mathcal{X}, W_{\mathcal{X}}, \lambda) = \bigg \Vert \sum_{(d, w) \in (D, W_D)} \nabla_{\theta}l_{rep}(\theta_t, d, w) \\
          &- \sum_{(\hat{d}, \hat{w}) \in (\mathcal{X} , W_{\mathcal{X}})} \nabla_{\theta}l_{rep}(\theta_t, \hat{d}, \hat{w}) \bigg \Vert^2 + \lambda \left \Vert W_{\mathcal{X}} \right \Vert^2\\
    \end{aligned}
\end{equation}
In the above equation, the replay loss function $l_{rep}$ is a weighted individual sample replay loss. In other words, {GradApprox} selects a subset of data points, and associated weights, such that the weighted sum of  individual samples' replay loss gradient is closest to the replay loss gradient of the entire dataset. Since this optimization is repeatedly applied after each task, the target of approximation  (gradient on $D$) also has weights $W_D$ learned from the previous round of GradApprox. Note that the use of weights is a \textit{necessary degree of freedom} in order to achieve the approximation; i.e., it allows us to feasibly approximate the overall gradient using only a subset of points. Further, these weights must necessarily be used in the subsequent learning algorithm, as the gradient approximation holds only under the conditions of weighted sum with chosen weights.


The above coreset optimization problem is generally applicable to any chosen replay loss function. For instance, using it on the \derpp~\cite{derpp} replay loss from Equation~\ref{eq:derloss} gives us a general  weighted loss: $l_{rep}(\theta, d, w) =  \alpha w {\left \Vert z - h_{\theta}(x) \right \Vert }^2 + \beta w l(y, f_{\theta}(x))$, where $d = (x, y, z)$, where the weights are selected by GradApprox. The first loss component is the distillation loss, and the second loss component is the label loss, each scaled by the respective hyperparameters $\alpha$ and $\beta$. Similarly, we can apply GradApprox to the vanilla Experience Replay loss (only cross-entropy on buffer) to improve upon the selected candidate pool for that loss function (\cite{doi:10.1080/09540099550039318}, also see Table~\ref{tab:er_gcr_table} in the supplementary materials). We apply GradApprox to the \gcrpp\ loss objective, described in the next section.



\noindent \textbf{GradApprox Implementation:}
Detailed pseudocode of the GradApprox algorithm is given in Algorithm~\ref{alg:gradapprox}. The optimization problem given in Eq~\eqref{subsetobj} is weakly submodular~\cite{killamsetty21a, Natarajan1995SparseAS}. Hence, we can effectively solve it using a greedy algorithm with approximation guarantees--we use Orthogonal matching pursuit\footnote{ A similar approach for subset selection using cross-entropy loss for efficient and robust supervised learning has been proposed in offline, single-task learning scenarios~\cite{killamsetty21a}.} to find the subset and their associated weights. We also added to Eq~\eqref{subsetobj} an $l_2$ regularization component over the weights of the replay buffer to discourage large weight assignments to data samples in the selected replay buffer, thereby preventing the model from overfitting on some samples. Finally, we select an equal number of samples from each class to make the selected replay buffer robust to the class imbalance in the dataset. In other words, if there are $C$ classes in the dataset $D$, we solve gradient approximation problems using per-class approximation by selecting a weighted subset of size $\frac{K}{C}$ from the data instances that pertain to the class being considered. 



\subsection{\gcrpp\ Loss objective}
\label{sec:loss_objective}
We adapt and modify the replay loss function from \derpp{}~\cite{derpp}  (current SOTA in offline continual learning) with the following two goals in mind: (a) to demonstrate the value of GradApprox as a replay buffer selection algorithm and (b) to achieve results, as an overall system, that beats current known SOTA performance in both offline and online continual learning. To use the replay loss from \derpp{}~\cite{derpp}, for each data sample $(x, y)$ in replay buffer and candidate pool, we also store historical model logit output $z$. Therefore, each data sample in the replay buffer and candidate pool consists of $(x, y, z)$.
%
 %
The formulation of the loss objective $\mathcal{L}(\theta)$ considered by \gcrpp\ is given below: 
\begin{equation}
\label{eq:gcrpobj}
\small
\begin{aligned}
    &\mathcal{L}(\theta) = \underbrace{\sum_{(x, y) \in D_{t}} l(y, f_{\theta}(x))}_{\text{(a)}} + \\
    &\underbrace{\underset{(x', y', z', w') \in (\mathcal{X}_{t-1} \doubleplus{} \mathcal{C}_t, W_{\mathcal{X}_{t-1}} \doubleplus{} W_{\mathcal{C}_t})}{\sum} \alpha w' {\left \Vert z' - h_{\theta}(x') \right \Vert }^2}_{\text{(b)}} + \\
    &\underbrace{\underset{(\hat{x}, \hat{y}, \hat{z}, \hat{w}) \in (\mathcal{X}_{t-1} \doubleplus{} \mathcal{C}_t, W_{\mathcal{X}_{t-1}} \doubleplus{} W_{\mathcal{C}_t})}{\sum} \beta \hat{w} l(\hat{y}, f_{\theta}(\hat{x}))}_{\text{(c)}} + \\
    &\underbrace{\underset{(\dot{x}, \dot{y}, \dot{z}, \dot{w}) \in (\mathcal{X}_{t-1} \doubleplus{} \mathcal{C}_t, W_{\mathcal{X}_{t-1}} \doubleplus{} W_{\mathcal{C}_t})}{\sum} \gamma \dot{w} l_{supcon}(\dot{x}, \dot{y}, \mathcal{X}_{t-1} \doubleplus{} \mathcal{C}_t, \theta)}_{\text{(d)}}
\end{aligned}
\end{equation}

where,
\begin{equation} 
     \small
     \begin{aligned}
     &l_{supcon}(x, y, \mathcal{A}, \theta) = \\
     &\frac{1}{|P(y)|}  \sum_{(\dot{x}, \dot{y}, \dot{z}) \in P(y)} \frac{\exp{}(\Omega_{\theta}(\dot{x}) \cdot \Omega_{\theta}(x))}{\underset{(\bar{x}, \bar{y}, \bar{z}) \in \mathcal{A}}{\sum}\exp{}(\Omega_{\theta}(\bar{x}) \cdot \Omega_{\theta}(x))} \\    
     &P(y) = \{(\dot{x}, \dot{y}, \dot{z}) \in \mathcal{A} : \dot{y} = y\} \\
     &\text{and }\doubleplus{} \text{ is a concatenation operator.}\\
     \end{aligned}
\end{equation}
The optimization objective consists of four components. The first component~(a) measures prediction loss compared to ground-truth labels for current task data. The second and third components~(b, c) measure the current model's loss computed over data from the combination of replay buffer and current task candidate pool, in terms of \textit{distillation loss} and label loss, respectively. As outlined in the previous section, \gcrpp\ uses a \textit{weighted loss} over the replay buffer, where the weights are an outcome of the replay buffer selection procedure (see previous sections). The fourth and final loss component~(d) is a weighted version of supervised contrastive loss~\cite{supcon}, initially proposed for standard supervised learning settings. This loss term improves the model's learned representations by forcing data samples from the same class to be closer in the embedding space than those from other classes. In our work, we always use  data samples from the combined candidate pool and replay buffer as anchor points for supervised contrastive learning loss.

\derpp\ also uses a loss criterion similar to (a,b,c) above. The primary differences in our loss function compared to \derpp\ are (1) the weighted-sum formulation of the loss, with weights selected by GradApprox optimization (our primary contribution), and (2) the use of supervised contrastive loss (d) to further improve the learned model (our secondary contribution). We show in ablation studies that these contributions independently and cumulatively add value over and above current SOTA. We also note that \gcrpp\ is applicable more broadly, to any specified replay loss funnctions (see e.g., ER~\cite{doi:10.1080/09540099550039318} based replay loss function, supplementary data~\ref{sec:generality}). 


\begin{algorithm}[!t]
\small{
\DontPrintSemicolon
\KwIn{Dataset: $D$, Parameters: $\theta$, Scalars: $\alpha$, $\beta$, $\gamma$, $\lambda$, Learning rate: $\eta$, Batch size:$B$, Buffer Size: $K$, Tolerance: $\epsilon$}
\KwOut{Parameters: $\theta$} 
\SetKwBlock{Begin}{function}{end function}
{
  Initialize Replay Buffer: $\mathcal{X} = \emptyset$, Replay Buffer weights: $W_{\mathcal{X}} = \emptyset $, and sample count: $n = 0$\;
    \For{$D_t \in D$ }
  {
    Initialize Candidate Pool: $\mathcal{C}_t = \emptyset$ and task sample count: $n_t = 0$ \;
    \For{$(x, y) \in D_t$}
    {
        Update task sample count: $n_t = n_t + 1$ and sample count: $n = n + 1$\;
        $\{(x', y', z', w')\} = \operatorname{Adaptive Sampling}(\mathcal{X}, \mathcal{C}_t, n_t, n)$ \;
        $x_{aug} = \operatorname{Augment}(x)$, $x'_{aug} = \operatorname{Augment}(x')$ \;
        Calculate model logit outputs: $z = h_{\theta}(x_{aug})$ \;
        $\theta = \theta - \eta \nabla_{\theta}\bigg(l(y, f_{\theta}(x_{aug})) + \alpha w' {\left \Vert z' - h_{\theta}(x'_{aug}) \right \Vert }^2 + \beta w' l(y', f_{\theta}(x'_{aug})) + \gamma w' l_{supcon}(x'_{aug}, y', \mathcal{X} \doubleplus{} \mathcal{C}, \theta)\bigg)$ \;
        $\mathcal{C} = \operatorname{Reservoir}(\mathcal{C}, (x, y, z), K)$ \;
    }
    $\mathcal{X}, W_{\mathcal{X}} = \operatorname{GradApprox}(\mathcal{X} \doubleplus{} \mathcal{C}_t, W \doubleplus{} \mathbbm{1}, \theta, \lambda, K, \epsilon)$
  }
}

\caption{\gcrpp Algorithm}\label{alg:gcr}
}
\end{algorithm}

\begin{algorithm}[!t]
\small{
\DontPrintSemicolon
\KwIn{Dataset: $D = \{(x_i, y_i, z_i)\}_{i=1}^{n}$, Existing Weights: $D_w$, Parameters: $\theta$, Scalar: $\lambda$, Budget: $K$, Tolerance: $\epsilon$}
\KwOut{Subset: $\mathcal{X}$, Subset Weights: $\mathcal{X}_w$} 
\SetKwBlock{Begin}{function}{end function}
{
  Initialize $Y = \operatorname{LabelCount}(D)$ \;
  Partition dataset $D$: $D = \{D_y\}_{y=1}^{Y}$ \;
  Partition dataset weights $W_D$ based on the labels of data samples: $W_D = \{W_{D_y}\}_{y=1}^{Y}$ \;
  Initialize Replay Buffer: $\mathcal{X} = \emptyset$ and Replay Buffer weights: $\mathcal{X}_w = \emptyset$ \;
  
  \For{$y \in \{1, 2, \cdots, Y\}$}
  {
    Initialize PerClass budget: $k_y = \frac{K}{Y}$, PerClass subset: $\mathcal{X}_y = \emptyset$ and PerClass weights: $W_{\mathcal{X}_{y}} = \mathbf{0}$ \;
    Calculate residuals : ${r} = \nabla_{\mathbf{w}} L_{sub}(D_y, D_{yw}, \mathcal{X}_{y}, \mathbf{w}, \lambda)|_{\mathbf{w} = \mathcal{X}_{yw}}$ \;
    \While{$|\mathcal{X}_y| \leq k_y$ and $L_{sub}(D_y, W_{D_y}, \mathcal{X}_{y}, W_{\mathcal{X}_{y}}, \lambda) \geq \epsilon$}
    {
        Find out maximum residual element: $e = \underset{j}{\text{argmax }} |r_j|$ \;
        Update PerClass subset: $\mathcal{X}_y = \mathcal{X}_y \cup \{e\}$ \;
        Calculate updated weights: $W_{\mathcal{X}_{y}} = \underset{\mathbf{w}}{\text{argmin }}L_{sub}(D_y, W_{D_{y}}, \mathcal{X}_{y}, \mathbf{w}, \lambda)$ \;
        Calculate residuals : ${r} = \nabla_{\mathbf{w}} L_{sub}(D_y, W_{D_{y}}, \mathcal{X}_{y}, \mathbf{w}, \lambda)|_{\mathbf{w} = W_{\mathcal{X}_{y}}}$ \;
    }
    Update subset: $\mathcal{X} = \mathcal{X} \doubleplus{} \mathcal{X}_{y}$ and subset weights: $W_{\mathcal{X}} = W_{\mathcal{X}} \doubleplus{} W_{\mathcal{X}_{y}}$
  }
}
\caption{GradApprox Algorithm}\label{alg:gradapprox}
}
\end{algorithm}


\noindent \textbf{Further optimization:} The candidate set for the current task is selected by reservoir sampling, specifically in order to maintain candidate data points \textit{alongside intermediate logits} for the distillation loss in the \gcrpp\ objective. It is possible to improve the candidate pool selection process further through a gradient-approximation objective at the end of current task training. However, this strategy would require maintaining intermediate logits for the entire dataset, requiring prohibitive storage requirements. 


\subsection{The \gcrpp\ algorithm}
We now put together the various components described above in a continual learning workflow. The pictorial representation of the \gcrpp\ algorithm was presented in Figure~\ref{fig:gcr_pipeline} and its detailed pseudocode in Algorithm~\ref{alg:gcr}. At each step \gcrpp\ trains the model on the current task data, and updates the replay buffer with a weighted summary of size $K$ selected from the candidate pool $\mathcal{C}_t$ and the previous replay buffer $\mathcal{X}_{t-1}$. 
\begin{equation}
\small
    \begin{aligned}
          \mathcal{X}_{t}, W_{\mathcal{X}_{t}} = \operatorname{GradApprox}(\mathcal{X}_{t-1} \doubleplus{} \mathcal{C}_t, W_{\mathcal{X}_{t-1}} \doubleplus{} W_{\mathcal{C}_t}, \theta, K) 
    \end{aligned}
\end{equation}
Each point in the candidate pool is initially assigned unit weight; i.e., $W_{\mathcal{C}_t} = \mathbbm{1}$. As shown in the pseudocode, we employ reservoir sampling~\cite{reservoir} for selecting the candidate pool. We also use adaptive sampling described in Algorithm~\ref{alg:adasampling} to sample tasks from the combined candidate pool and the replay buffer. Due to space constraints, we give detailed pseudocode of the Adaptive Sampling Algorithm in Supplementary materials~\ref{sec:algorithm}. Finally, the pseudocode given in Algorithm~\ref{alg:gcr} requires the knowledge of task boundaries to update replay buffer using GradApprox; however, we can also use streaming boundaries or regular intervals of data samples as boundaries in practice to implement the $\gcrpp$ algorithm, thereby making it task agnostic.   

\section{Experiment setup}
We test the efficacy of our approach \gcrpp\ by comparing its performance with different state-of-the-art continual learning baselines under different CL settings. 

\begin{table*}[ht]
\fontsize{9pt}{10pt} \selectfont
\centering
\scalebox{0.8}{
\begin{tabular}{c|l|ccc|ccc|ccc}
       & & \multicolumn{3}{c|}{\textbf{S-Cifar-10}} & \multicolumn{3}{c|}{\textbf{S-Cifar-100}} & \multicolumn{3}{c}{\textbf{S-TinyImageNet}} \\                    
Setting & Method & K=200               & K=500               & K=2000              & K=200              & K=500              & K=2000              & K=200               & K=500               & K=2000     \\ \midrule
\multirow{ 6}{*}{Class-IL} & ER     & 49.16±2.08          & 62.03±1.70          & 77.13±0.87          & 21.78±0.48         & 27.66±0.61         & 42.80±0.49          & 8.65±0.16           & 10.05±0.28          & 18.19±0.47 \\
& GEM    & 29.99±3.92          & 29.45±5.64          & 27.2±4.5            & 20.75±0.66         & 25.54±0.65         & 37.56±0.87          & -                   & -                   & -          \\
 & GSS    & 38.62±3.59          & 48.97±3.25          & 60.40±4.92          & 19.42±0.29         & 21.92±0.34         & 27.07±0.25          & 8.57±0.13           & 9.63±0.14           & 11.94±0.17 \\
& iCARL  & 32.44±0.93          & 34.95±1.23          & 33.57±1.65          & 28.0±0.91          & 33.25±1.25         & 42.19±2.42          & 5.5±0.52            & 11.0±0.55           & 18.1±1.13  \\
& DER    & 63.69±2.35          & 72.15±1.31          & 81.00±0.97          & 31.23±1.38         & 41.36±1.76         & 55.45±0.86          & \textbf{13.22±0.92} & \textbf{19.05±1.32}          & 31.53±0.87 \\ 
& \textbf{\gcrpp}  & 64.84±1.63          & \textbf{74.69±0.85} & \textbf{83.97±0.58} & \textbf{33.69±1.4} & \textbf{45.91±1.3} & \textbf{60.09±0.72} & \textbf{13.05±0.91}          & \textbf{19.66±0.68} & \textbf{35.68±0.52}

\\ \midrule
\multirow{ 6}{*}{Task-IL} & ER     & 91.92±1.01          & 93.82±0.41          & 96.01±0.28          & 60.19±1.01          & 66.23±1.52          & 74.67±1.2           & 38.83±1.15         & 47.86±0.87          & 62.04±0.7          \\
& GEM    & 88.67±1.76          & 92.33±0.8           & 94.34±1.31          & 58.84±1.0           & 66.31±0.86          & 74.93±0.6           & -                  & -                   & -                  \\
& GSS    & 90.0±1.58           & 91.73±1.18          & 93.54±1.32          & 55.38±1.34          & 60.28±1.18          & 66.88±0.88          & 31.77±1.34         & 36.52±0.91          & 43.75±0.83         \\
& iCARL  & 74.59±1.24          & 75.63±1.42          & 76.97±1.04          & 51.43±1.47          & 58.16±1.76          & 67.95±2.69          & 22.89±1.83         & 35.86±1.07          & 49.07±2.0          \\
& DER    & \textbf{91.91±0.51} & 93.96±0.37          & 95.43±0.26          & 63.09±1.09          & \textbf{71.73±0.74}          & 78.73±0.61          & \textbf{42.27±0.9} & \textbf{53.32±0.92} & 64.86±0.48         \\ 
& \textbf{\gcrpp}  & 90.8±1.05           & \textbf{94.44±0.32} & \textbf{96.32±0.18} & \textbf{64.24±0.83} & \textbf{71.64±2.1}           & \textbf{80.22±0.49} & \textbf{42.11±1.01}         & \textbf{52.99±0.89}          & \textbf{66.7±0.59} \\ \hline
\end{tabular}}
\caption{Offline Class-IL and Task-IL Continual Learning. See text for details.  Numbers represent mean ± SEM of model test accuracy over 15 runs. Best-performing models in each column are bolded (paired $t$-test, $p<0.05$). Subsequent tables follow the same style.   }\label{tab:offline_table}
\end{table*}
\begin{table*}[ht]
\fontsize{9pt}{10pt} \selectfont
\centering
\scalebox{0.8}{
\begin{tabular}{c|l|ccc|ccc|ccc}
       & & \multicolumn{3}{c|}{\textbf{S-Cifar-10}} & \multicolumn{3}{c|}{\textbf{S-Cifar-100}} & \multicolumn{3}{c}{\textbf{S-TinyImageNet}} \\
Setting & Method & K=200               & K=500               & K=2000             & K=200              & K=500              & K=2000              & K=200               & K=500              & K=2000              \\ \midrule
\multirow{ 2}{*}{Class-IL} & DER    & 35.79±2.59          & 24.02±1.63          & 12.92±1.1          & 62.72±2.69         & 49.07±2.54         & 28.18±1.93          & 64.83±1.48          & 59.95±2.31         & 39.83±1.15          \\ 
& \textbf{\gcrpp}  & \textbf{32.75±2.67}          & \textbf{19.27±1.48} & \textbf{8.23±1.02} & \textbf{57.65±2.48}         & \textbf{39.2±2.84} & \textbf{19.29±1.83} & 65.29±1.73          & \textbf{56.4±1.08}          & \textbf{32.45±1.79} 
\\ \midrule
\multirow{ 2}{*}{Task-IL} & DER    & 6.08±0.7   & 3.72±0.55  & 1.95±0.32  & 25.98±1.55          & 25.98±1.55          & 7.37±0.85          & 40.43±1.05 & 28.21±0.97          & 15.08±0.49         \\
& \textbf{\gcrpp}  & 7.38±1.02  & \textbf{3.14±0.36}  & \textbf{1.24±0.27}  & \textbf{24.12±1.17} & \textbf{15.07±1.88} & \textbf{5.75±0.72} & 40.36±1.08 & \textbf{27.88±1.19} & \textbf{13.1±0.57}
\\ \hline
\end{tabular}}
\caption{Forgetting metric (lower is better) in Offline Class-IL and Task-IL Continual Learning. For simplicity, we only present numbers for the two best-performing algorithms--DER and GCR (ours). Full table in appendix. }\label{tab:forgetting_offline_trim}
\end{table*}

\subsection{Continual Learning settings}
We compare the performance of \gcrpp\ with the baselines considered in the following continual learning settings.\\
\noindent \textbf{Offline Class-Incremental (Class-IL)}--All data for the current task is available for training with multiple learning iterations. Previous tasks' data are only partially available through the replay buffer.
\noindent \textbf{Offline Task-Incremental (Task-IL)} uses different output heads/ classifier layers for different tasks, and needs a task identifier at inference time to select the appropriate classifier. \noindent \textbf{Online Streaming}--Data arrives in the form of long streams; the learner can iterate only once on  streaming data and multiple times on the stored replay buffer.

\subsection{Baselines}
We consider the following continual learning methods as baselines for a thorough evaluation of \gcrpp\ performance.
\noindent \textbf{ER}-Experience replay, a rehearsal method with random sampling in Memory-Retrieval and reservoir sampling in Memory-Update.
\noindent \textbf{GEM}-Gradient Episodic Memory, focus on minimizing negative backward transfer (catastrophic forgetting) by the efficient use of episodic memory.
\noindent \textbf{MIR}-Maximally Interfered Retrieval, which retrieves memory samples suffering from an increase in loss given estimated parameters update for the current task.
\noindent \textbf{GSS}-Gradient-Based Sample Selection, which diversifies the gradients of the samples in the replay memory.
\noindent \textbf{iCaRL}-incremental classifier and representation learning, which simultaneously learns classifiers and feature representation in the class-incremental setting.
\noindent \textbf{\derpp}-Dark Experience Replay, a strong baseline built upon Reservoir Sampling that matches the network’s logits sampled throughout the optimization trajectory, thus promoting consistency with its past. We compare against ER, GEM, GSS, iCaRL, and \derpp baselines in the offline CL setting. Whereas for the online streaming setting, we compare with MIR in addition to all the above mentioned baselines.  



\subsection{Datasets}
We work on the following datasets: \emph{Sequential CIFAR-10}, which splits the CIFAR-10 dataset \cite{Krizhevsky09learningmultiple} into 5 different tasks with non-overlapping classes and 2 classes in each task; \emph{Sequential CIFAR-100}, obtained similarly by Cifar-100 \cite{Krizhevsky09learningmultiple} into 5 tasks of 10 classes each, \emph{Sequential Tiny-ImageNet} which splits Tiny-ImageNet \cite{Pouransari2014TinyIV} into 10 tasks of 20 classes each, and \emph{Sequential ImageNet-1k} which splits ImageNet 1k~\cite{ILSVRC15} into 5 tasks of 200 classes each. For an extended learning experiment, we also split Cifar-100 into 20 tasks of 5 classes each (see Results below). 


\subsection{Model architecture and other training details}
We used ResNet18 as the base model architecture for all datasets throughout all experiments. We use a minibatch size of 32 for both the current task data and the replay buffer. We use a standard stochastic gradient descent (SGD) optimizer without any learning rate scheduler. Other parameter values that are not explicitly mentioned are hyperparameters, and we calculate the best performing values for each run using hyperparameter tuning. Details of grid search and selected values are added in the supplementary material.


\subsection{Evaluation}
Following previous work~\cite{derpp}, we select hyperparameters by performing a grid search on validation data obtained by splitting 10\% of training data. In all comparisons, the respective models are trained with chosen hyperparameters on the training data, and model accuracy is evaluated on a separate test dataset. This procedure is repeated 15 times with different random seeds controlling weight initialization and subsequent data randomization. We present the mean accuracy along with the standard error of the mean. We also present the \textit{forgetting metric} (F)~\cite{chaudhry2018riemannian}, which measures how much the accuracy of  learnt  tasks  decrease over time  as  the  continual  models  tries  to learn subsequent tasks. Since all models within a comparison are trained on the same set of random seeds, we use paired $t$-tests to determine the statistical significance of any differences in metrics, evaluated at $p < 0.05$ confidence threshold.

\section{Results}
\subsection{Offline Continual Learning}
\label{offline-cl}

\begin{table*}[ht]
\fontsize{9pt}{10pt} \selectfont
\centering
\scalebox{0.8}{
\begin{tabular}{l|cccc|cccc}
       & \multicolumn{4}{c|}{\textbf{Class-IL}}                                             & \multicolumn{4}{c}{\textbf{Task-IL}}                                              \\
       & \multicolumn{2}{c|}{K=200}                   & \multicolumn{2}{c|}{K=500} & \multicolumn{2}{c|}{K=200}                   & \multicolumn{2}{c}{K=500} \\
Method & Acc ($\uparrow$)        & \multicolumn{1}{c|}{F ($\downarrow$)}          & Acc ($\uparrow$)         & F ($\downarrow$)          & Acc ($\uparrow$)       & \multicolumn{1}{c|}{F ($\downarrow$)}         & Acc ($\uparrow$)        & F ($\downarrow$)          \\ \midrule
ER     & 9.57±3.06  & \multicolumn{1}{c|}{76.08±4.38} & 15.79±4.63   & 69.46±5.26  & 68.66±8.37 & \multicolumn{1}{c|}{19.84±7.72} & 75.75±4.94  & 14.18±4.62  \\
DER  & 15.64±2.59 & \multicolumn{1}{c|}{64.34±4.33} & 26.61±4.94   & 51.27±7.57  & 66.87±3.91 & \multicolumn{1}{c|}{21.08±3.68} & 77.41±4.15  & 13.05±4.17  \\
\textbf{\gcrpp}    & \textbf{21.02±1.99} & \multicolumn{1}{c|}{\textbf{51.82±4.04}} & \textbf{31.93±4.86}   & \textbf{42.1±6.6}    & \textbf{68.86±2.98} & \multicolumn{1}{c|}{20.11±3.07} & 78.09±3.33  & 12.97±3.15 
\end{tabular}}
\caption{Offline Class-IL and Task-IL results on S-Cifar100 (20 Tasks) for 20 tasks with each task having 5 classes. GCR significantly outperforms ER \& DER, by larger margins than the 5-task setting, showing that GCR coreset selection scales effectively to more tasks. }\label{tab:offline_more_tasks_table}
\end{table*}
\begin{table*}[!ht]
\fontsize{9pt}{10pt} \selectfont
\centering
\scalebox{0.8}{
\begin{tabular}{l|ccc|ccc|ccc}
       & \multicolumn{3}{c|}{\textbf{S-Cifar-10}}                     & \multicolumn{3}{c|}{\textbf{S-Cifar-100}}                      & \multicolumn{3}{c}{\textbf{S-TinyImageNet}}                  \\
Method & K=200              & K=500              & K=2000             & K=200               & K=500              & K=2000              & K=200              & K=500             & K=2000              \\ \midrule
ER     & 32.63±5.11         & 43.32±4.75         & \textbf{49.31±7.45}         & 11.72±1.25          & 14.94±1.69         & 20.78±1.48          & 4.36±1.02          & 6.69±0.53         & 11.3±1.45           \\
GEM    & 16.81±1.17         & 17.73±0.98         & 20.35±2.44         & 10.42±1.66          & 9.18±1.8           & 12.38±3.81          & 5.15±0.71          & 6.14±1.01         & 9.56±2.63           \\
GSS    & 33.83±3.95         & 40.59±4.2          & 41.7±4.56          & 12.31±0.57          & 13.84±0.57         & 16.24±0.8           & 5.82±0.38          & 6.78±0.41         & 8.44±0.48           \\
iCARL  & 38.66±2.03         & 33.53±3.17         & 39.56±1.53         & 11.23±0.31          & 11.81±0.29         & 12.28±0.39          & 4.39±0.19          & 5.36±0.27         & 6.66±0.3            \\
MIR    & 21.5±0.7           & 30.52±1.22         & 46.40±1.81         & 10.3±0.3            & 11.40±0.38         & 16.02±0.5           & 4.9±0.17           & 5.1±0.33          & 7.05±0.3            \\
DER    & 38.26±5.08         & 45.56±6.02         & \textbf{49.9±7.55}          & 10.68±2.03          & 14.43±2.44         & \textbf{24.83±1.58} & 4.81±0.57          & \textbf{8.49±0.48}         & \textbf{15.97±0.95} \\
\textbf{\gcrpp}  & \textbf{42.26±7.6} & \textbf{51.17±3.5} & \textbf{52.3±7.01} & \textbf{14.88±0.94} & \textbf{18.89±1.5} & 22.47±2.01          & \textbf{6.19±0.36} & \textbf{8.77±0.5} & 11.83±1.26         
\end{tabular}
}
\caption{Online Continual Learning. See text for details.    }
\label{tab:online_table}
\end{table*}
\begin{table}[h]
\fontsize{9pt}{10pt} \selectfont
\centering
\scalebox{0.8}{
\begin{tabular}{l|cc}

Ablation Setting $\backslash$ Buffer Size                                               & 500                & 2000                \\ \hline
GCR (ours)                           & 45.91±1.3 & 60.09±0.72 \\
GCR w/ reservoir sampling                        & 43.73±1.7          & 57.41±1.01          \\ 
GCR w/ bilevel coresets                      & 40.41±1.05         & 51.11±0.96          \\
DER++                                &  41.36±1.76        &     55.45±0.86 \\  \midrule \midrule
GCR (ours)                               & 45.91±1.3          & 60.09±0.72 \\
GCR w/o SupCon Loss                                & 45.54±1.39         & 58.35±0.78    \\    \midrule \midrule

\end{tabular}}
\caption{Ablation Study on Class-IL offline continual learning with S-Cifar-100 for two different buffer sizes. For buffer size=2000, all pairwise differences were statistically significant with $p< 1e-3$. }\label{tab:ablation_table1}
\end{table}
Table \ref{tab:offline_table} shows the average accuracy of various continual learning methods at the end of tasks for Cifar-10, Cifar-100 and Tiny-ImageNet for different buffer sizes. Models are trained for 50 epochs for Cifar-10 and Cifar-100. Epochs are increased to 100 for Tiny-ImageNet, which is commonly done for harder datasets. The candidate pool size in \gcrpp\ is kept equal to the buffer size for all experiments. 

In the Class-IL setting, \gcrpp\ outperforms other methods with gains of roughly 1-5\% that are statistically significant. For S-Cifar-100, for buffer sizes of 500  and 2000, the difference is more than 4.5\%. Among the baselines considered, only \derpp\ appears close; this can be attributed to the usage of distillation loss proposed by \derpp, which other methods do not use. S-TinyImageNet is a challenging dataset for continual learning,  and both \derpp\ and \gcrpp\ show significant improvements over alternative approaches.

For the Task-IL setting, although \gcrpp\  outperforms others in many instances (e.g., in S-Cifar-100 for buffer size 2000, the difference is close to 1.5\%),  \derpp and ER both have accuracy reasonably close to that of \gcrpp. We believe this is due to the substantially lower complexity of the Task-IL setting; the CL model only needs to learn each task in isolation. GSS gives comparable results on S-Cifar-10, but its performance deteriorates for S-Cifar-100 and S-TinyImageNet, suggesting that it does not scale to more difficult datasets.

\noindent \textbf{Forgetting metric:} Table~\ref{tab:forgetting_offline_trim} shows the forgetting metric (F) for different settings and buffer sizes. This metric shows that \gcrpp\ performs better due to less catastrophic forgetting than other replay based methods; the forgetting metric and accuracy numbers in Table \ref{tab:offline_table} covary closely. 

\noindent \textbf{Increased number of tasks:} Table~\ref{tab:offline_more_tasks_table} shows accuracy and forgetting metric on the S-Cifar-100 20-task data (5 classes per task) to study the impact of increasing numbers of rounds of gradient approximation / replay buffer updates. \gcrpp\ shows robust gains, especially at smaller buffer sizes and in Class-incremental setting, which correspond to the significantly harder continual learning problems. The wider gap between \derpp and \gcrpp\ compared to Table~\ref{tab:offline_table} shows that our coreset selection procedure accumulates value over time, scaling effectively to 4x the number of tasks.

\noindent \textbf{Gradient-based selection on high-res images:} Finally, we show that the gradient optimization works for significantly more complex tasks or input characteristics, by demonstrating gains in the high-resolution S-ImageNet-1k task (see Supplementary Table~\ref{tab:imagenet}).

\noindent \textbf{Generality of gradient-based coresets:}
We also show that our gradient based coresets can be combined effectively with other replay based approaches to improve their performance. Refer Supplementary materials \ref{sec:generality} for details. 

\subsection{Online Streaming Continual Learning}
\label{online-cl}

Table \ref{tab:online_table} compares \gcrpp\  against baselines in the online setting. Here, data is made available in streams of size 6250, and the task boundary is blurred. Due to the streaming nature of the data, models are only allowed a single epoch of training on each stream.  Since iCARL and GEM require knowledge of task boundaries, we explicitly provided the task identity while training those two baselines for a fair comparison. In this setting, the performance gap between \gcrpp\ and baselines is much more apparent for low buffer sizes, where the quality of examples stored plays a significant role. As buffer size increases, coverage can compensate for suboptimal data selection. In fact, for S-Cifar-10 with a buffer size of 500, the gap between \derpp\ and \gcrpp\ is more than 5\%. Forgetting metrics are reported in Supplementary materials~\ref{sec:moreresults} due to space considerations. Recent papers like \cite{shim2021online} in online continual learning do not compare against techniques in an offline setting that can be straightforwardly adapted and are very competitive in the online setting. We note that their reported numbers are significantly lower even than \derpp in Table~\ref{tab:online_table}; as a result, we did not re-evaluate those techniques in our study.


\subsection{Ablation studies}
Table~\ref{tab:ablation_table1} shows the performance of \gcrpp\ with the coreset selection method being replaced by reservoir sampling~\cite{derpp} and bilevel coresets~\cite{borsos2020coresets} (rows 2, 3 respectively). Shown for comparison is \derpp, which uses reservoir sampling, and also does not have the supervised contrastive loss used in our \gcrpp\ formulation. We see that bilevel coresets performs worse than \derpp, and worse even than reservoir sampling, and \gcrpp\ clearly outperforms. In similar vein, rows 5, 6 compare \gcrpp\ with and without supervised contrastive loss, showing that SupCon contributes clear, additional value. All pairwise comparisons for buffer size 2000 were statistically significant  with $p< 1e-3$. 




\section{Conclusion, Limitations, and Future Work}
We presented \gcrpp, a gradient-based coreset selection method for replay-based continual learning in which we propose {gradient approximation} as an optimization criterion for selecting coresets, building on recent advances in supervised learning settings~\cite{killamsetty21a}. We integrate this objective into the continual learning workflow in selecting and updating replay buffers for future training. We also include a supervised representation learning loss~\cite{supcon} in our CL objective, enhancing the learned representations over the model's lifetime. Extensive experiments across datasets, replay buffer sizes, and CL settings (offline/online, class-IL/task-IL) show that \gcrpp\ significantly outperforms previous approaches in comparable settings, to a degree of 2-4\% accuracy in offline settings and up to 5\% in online settings. Ablation studies show that the \gcrpp\ coreset selection objective outperforms previous best selection mechanisms~\cite{borsos2020coresets,derpp}, and the representation loss also independently contributes to performance gains. Our coreset selection approach scales well, providing increasingly significant gains as the number of tasks increases, according to experiments with a high number of tasks and image complexity. We demonstrate that the core ideas of \gcrpp\ apply to offline and online settings (typically studied separately) by combining them. We hope that further cross-fertilization of ideas will continue to occur. A challenge remains, however, in integrating \gcrpp\ with previous ideas in the field (e.g., maintaining exemplars (iCaRL~\cite{rebuffi2017icarl}), function regularization~\cite{riemer2018learning}, and self-supervision). In addition, we simplify our candidate selection by using a throw-away candidate buffer, which adds memory overhead that may be problematic in online CL. In the future, we will work on developing streaming coreset selection mechanisms to address these challenges.

\section{Acknowledgments and  Funding}
We thank the CVPR area chairs and anonymous reviewers for their constructive comments on this paper. RI and KK were supported by the National Science Foundation(NSF) under Grant Number 2106937, a startup grant from UT Dallas, as well as Google and Adobe awards. The opinions, findings, conclusions, and recommendations expressed in this material are those of the author(s). They do not necessarily reflect the views of the National Science Foundation, Google, and Adobe.

\bibliographystyle{unsrt}  
\bibliography{gcr}  
\appendix
\section{Algorithm}\label{sec:algorithm}
\begin{algorithm}[!tbh]
\small{
\DontPrintSemicolon
\KwIn{Replay Buffer: $\mathcal{X}$, Replay Buffer weights: $W$, Candidate Pool: $\mathcal{C}_t$, Task sample count: $n_t$, Entire sample count: $n$}
\KwOut{Data sample: \{(x, y, z, w)\}} 
\SetKwBlock{Begin}{function}{end function}
{
  Calculate probability: $p = \frac{n_t}{n}$ \;
  Sample a random number: $p_f = random([0, 1])$ \;
  \If{$p_f \le p$}
    {
    Sample a index randomly: $i = random([1, |\mathcal{C}_t|])$ \;
    $(x, y, z, w) = (\mathcal{C}_{t}[i], 1)$
    }
  \Else{
    Sample a index randomly: $i = random([1, |\mathcal{X}|])$ \;
    $(x, y, z, w) = (\mathcal{X}[i], W[i])$
  }
}
\caption{Adaptive Sampling Algorithm}\label{alg:adasampling}
}
\end{algorithm}

\section{Additional Results}\label{sec:moreresults}
\subsection{Forgetting metric}
\begin{table*}[ht]
\fontsize{9pt}{10pt} \selectfont
\centering
\scalebox{0.8}{
\begin{tabular}{c|l|ccc|ccc|ccc}
       & & \multicolumn{3}{c|}{\textbf{S-Cifar-10}} & \multicolumn{3}{c|}{\textbf{S-Cifar-100}} & \multicolumn{3}{c}{\textbf{S-TinyImageNet}} \\
Setting & Method & K=200               & K=500               & K=2000             & K=200              & K=500              & K=2000              & K=200               & K=500              & K=2000              \\ \midrule
& ER     & 59.3±2.48           & 43.22±2.1           & 23.85±1.09         & 75.06±0.63         & 67.96±0.78         & 49.12±0.57          & 76.53±0.51          & 75.21±0.54         & 65.58±0.53          \\
& GEM    & 80.36±5.25          & 78.93±6.53          & 82.33±5.83         & 77.4±1.09          & 71.34±0.78         & 55.27±1.37          & -                   & -                  & -                   \\
Class-IL & GSS    & 72.48±4.45          & 59.18±4.0           & 44.59±6.13         & 77.62±0.76         & 74.12±0.42         & 67.42±0.62          & 76.47±0.4           & 75.3±0.26          & 72.49±0.43          \\
& iCARL  & \textbf{23.52±1.27} & 28.2±2.41           & 21.91±1.15         & \textbf{47.2±1.23} & 40.99±1.02         & 30.64±1.85          & \textbf{31.06±1.91} & \textbf{37.3±1.42} & 39.88±1.51          \\
& DER    & 35.79±2.59          & 24.02±1.63          & 12.92±1.1          & 62.72±2.69         & 49.07±2.54         & 28.18±1.93          & 64.83±1.48          & 59.95±2.31         & 39.83±1.15          \\ 
& \textbf{\gcrpp}  & 32.75±2.67          & \textbf{19.27±1.48} & \textbf{8.23±1.02} & 57.65±2.48         & \textbf{39.2±2.84} & \textbf{19.29±1.83} & 65.29±1.73          & 56.4±1.08          & \textbf{32.45±1.79} 
\\ \midrule
& ER     & 6.07±1.09  & 3.5±0.53   & 1.37±0.44  & 27.38±1.46          & 17.37±1.06          & 8.03±0.66          & 40.47±1.54 & 30.73±0.62          & 18.0±0.83          \\
& GEM    & 9.57±2.05  & 5.6±0.96   & 2.95±0.81  & 29.59±1.66          & 20.44±1.13          & 9.5±0.73           & -          & -                   & -                  \\
Task-IL & GSS    & 8.49±2.05  & 6.37±1.55  & 4.31±1.68  & 32.81±1.75          & 26.57±1.34          & 18.98±1.13         & 50.75±1.63 & 45.59±0.99          & 38.05±1.17         \\
& iCARL  & 25.34±1.64 & 22.61±3.97 & 24.47±1.36 & 36.2±1.85           & 27.9±1.37           & 16.99±1.76         & 42.47±2.47 & 39.44±0.84          & 30.45±2.18         \\
& DER    & 6.08±0.7   & 3.72±0.55  & 1.95±0.32  & 25.98±1.55          & 25.98±1.55          & 7.37±0.85          & 40.43±1.05 & 28.21±0.97          & 15.08±0.49         \\
& \textbf{\gcrpp}  & 7.38±1.02  & \textbf{3.14±0.36}  & \textbf{1.24±0.27}  & \textbf{24.12±1.17} & \textbf{15.07±1.88} & \textbf{5.75±0.72} & 40.36±1.08 & \textbf{27.88±1.19} & \textbf{13.1±0.57}
\\ \hline
\end{tabular}}
\caption{Forgetting metric in Offline Class-IL and Task-IL Continual Learning}\label{tab:forgetting_offline_full}
\end{table*}
\begin{table*}[!ht]
\fontsize{9pt}{10pt} \selectfont
\centering
\scalebox{0.8}{
\begin{tabular}{l|ccc|ccc|ccc}
       & \multicolumn{3}{c|}{\textbf{S-Cifar-10}}                     & \multicolumn{3}{c|}{\textbf{S-Cifar-100}}                      & \multicolumn{3}{c}{\textbf{S-TinyImageNet}}                  \\
Method & K=200      & K=500              & K=2000              & K=200              & K=500              & K=2000            & K=200             & K=500              & K=2000             \\ \hline
ER     & 47.01±6.63 & 38.72±7.94         & 31.96±8.93          & 30.16±0.69         & 26.29±1.31         & 16.42±2.17        & 27.86±1.69        & 32.53±1.18         & 27.91±1.41         \\
GEM    & 73.63±3.96 & 73.07±6.58         & 53.27±10.93         & 32.94±2.88         & 27.15±3.78         & 29.97±7.12        & -                 & -                  & -                  \\
GSS    & 48.8±6.56  & 40.62±6.74         & 40.67±5.75          & 33.06±1.05         & 25.37±1.93         & 19.56±1.64        & 36.91±1.44        & 32.67±1.36         & 23.63±1.18         \\
iCARL  & 23.78±3.64 & 26.2±4.31          & \textbf{22.11±4.61} & \textbf{9.53±0.57} & \textbf{9.15±0.49} & \textbf{8.9±0.49} & \textbf{6.95±0.5} & \textbf{7.22±0.38} & \textbf{6.89±0.37} \\
DER    & 34.12±7.04 & 29.05±8.59         & 27.5±8.69           & 26.84±1.7          & 22.92±2.73         & 13.72±2.03        & 31.68±1.46        & 27.09±0.79         & 14.97±1.28         \\
\gcrpp\  & 26.7±8.37  & \textbf{20.1±3.32} & \textbf{22.18±9.9}  & 21.86±1.77         & 19.46±1.72         & 17.91±2.3         & 34.19±1.07        & 27.47±0.8          & 22.31±1.35        
\end{tabular}
}
\caption{Forgetting metric in Online Continual Learning.}
\label{tab:online_forgetting_table}
\end{table*}
In this section we present additional results for the experiments shown in Section \ref{offline-cl} and \ref{online-cl}. We report the \textit{forgetting} metric (FRG) which shows how much the accuracy of learnt tasks over time as the continual models tries to learn subsequent tasks. The average forgetting across the tasks is reported in Table~\ref{tab:forgetting_offline_full} (offline setting), and Table~\ref{tab:online_forgetting_table} (online streaming setting). 
It is worth noting that FRG should only be seen along with final accuracy to draw comparisons between two continual learning  models. This is because FRG alone can be misleading--a model which does not learn any subsequent tasks throughout the training will give near to 0 forgetting but will give random final accuracy which is undesirable. A clear example is that of iCaRL~\cite{rebuffi2017icarl} in the offline setting; we see that the method has poor overall accuracy (Table~\ref{tab:offline_table}) but highly favorable forgetting metrics (Table~\ref{tab:forgetting_offline_full}).

\subsection{OCS vs GCR}
\begin{table*}[ht]
\fontsize{9pt}{10pt} \selectfont
\centering
\scalebox{0.8}{
\begin{tabular}{l|c}
       & \textbf{OCS vs GCR} \\
Method & S-Cifar 100 (20 Tasks)         \\ \hline
OCS    & 60.5±0.55           \\
GCR    & 60.86±3.53 
\end{tabular}}
\caption{Comparing OCS with GCR for Task-IL setting of S-Cifar-100 (20 tasks) and buffer size of 100.}\label{tab:ocs_gcr_table}
\end{table*}
Table \ref{tab:ocs_gcr_table} compares published performance numbers from one setting, for the OCS algorithm~\cite{ocs}, against our trained model in the same setting. Our approach shows better performance in the comparison, suggesting that our gradient approximation objective is superior to the gradient diversity-based selection objective of OCS. However, this comparison is incomplete; the authors of OCS have not made their code available for comparison, the paper's description of the algorithm was insufficient for reproduction, and they did not publish numbers on any of the other settings, datasets, buffer sizes we explored in our paper.

\section{Generality of gradient-based coresets}
\label{sec:generality}
\begin{table*}[ht]
\fontsize{9pt}{10pt} \selectfont
\centering
\scalebox{0.8}{
\begin{tabular}{l|llll|llll}
       & \multicolumn{4}{c|}{\textbf{S-Cifar-10}}                                             & \multicolumn{4}{c}{\textbf{S-Cifar-100}}                                            \\
       & \multicolumn{2}{c|}{\textbf{Class-IL}}                & \multicolumn{2}{c|}{\textbf{Task-IL}} & \multicolumn{2}{c|}{\textbf{Class-IL}}                & \multicolumn{2}{c}{\textbf{Task-IL}} \\
Method & K=500      & \multicolumn{1}{l|}{K=2000}     & K=500         & K=2000       & K=500      & \multicolumn{1}{l|}{K=2000}     & K=500         & K=2000      \\ \hline
ER     & 62.03±1.70 & \multicolumn{1}{l|}{77.13±0.87} & 93.82±0.41    & 96.01±0.28   & 27.66±0.61 & \multicolumn{1}{l|}{42.80±0.49} & 66.23±1.52    & 74.67±1.2   \\
ER+\gcrpp & \textbf{66.66±2.1}  & \multicolumn{1}{l|}{\textbf{80.15±1.17}} & 94.17±0.46    & \textbf{96.47±0.22}   & \textbf{30.68±0.47} & \multicolumn{1}{l|}{\textbf{47.09±1.08}} & \textbf{70.25±0.81}    & \textbf{78.59±0.5}  
\end{tabular}}
\caption{GCR coreset selection with ER method. Numbers represent mean ± SEM of model test accuracy over 15 runs. Best-performing models in each column are bolded (paired $t$-test, $p<0.05$). }\label{tab:er_gcr_table}
\end{table*}
We also examined whether the gains from our gradient approximation procedure for coreset selection (Section~\ref{sec:buffer_selection}) were dependent on the specific loss function that we use for CL (Section~\ref{sec:loss_objective}). To evaluate this, we enhanced ER~\cite{riemer2018learning} (a simple replay-based continual learning procedure that does not use the distillation loss from Section~\ref{sec:loss_objective}) with our gradient approximation procedure. The results in Table~\ref{tab:er_gcr_table} show that the gains from GCR are robust, and apply to other replay-based methods as well. Other baseline methods like iCARL, GSS, etc have specific replay buffer selection methods, unlike ER which uses random samples, and it was not clear how to add GCR on top of those methods. In any case, our results show that GCR beats those methods by significant margins.

\begin{table*}[ht]
\fontsize{9pt}{10pt} \selectfont
\centering
\scalebox{0.8}{
\begin{tabular}{l|cc}
\multicolumn{1}{c|}{Method} & Class-IL                    & Task-IL                     \\ \hline
ER [36]                         & 7.43 &  15.32 \\
DER++ [6]                        & 10.22                       & 17.79                       \\
GCR                         & \textbf{11.33}              & \textbf{19.03}             
\end{tabular}}
\caption{Scaling up to S-ImageNet1k (5 tasks, buffer size 1000)}
\label{tab:imagenet}
\end{table*}

Finally, we conducted experiments on the significantly more difficult S-Imagenet-1k dataset~\cite{ILSVRC15}, comprising high-resolution Imagenet images broken down into 5 tasks of 200 categories each. Table~\ref{tab:imagenet} shows that \gcrpp\ outperforms ER and \derpp\ by significant margin. Note, however, that  all three methods have fairly low accuracy on the task overall; this is expected given that the task is significantly harder than S-Cifar100.


\section{Implementation details}

\subsection{Hyperparameter Search}
\begin{table*}[ht]
\fontsize{9pt}{10pt} \selectfont
\centering
\scalebox{0.8}{
\begin{tabular}{|l|c|l|l|l|}
\multicolumn{5}{|c|}{\textbf{Offline Class-IL}}                                                                                                                                                                                                                                                                                                                                                                                                                                                                                                              \\
\multicolumn{1}{|c|}{Method} & Buffer Size                                              & \multicolumn{1}{c|}{S-Cifar10}                                                                                                                       & \multicolumn{1}{c|}{S-Cifar100}                                                                                                                      & \multicolumn{1}{c|}{S-Tinyimg}                                                                                                                       \\ \hline
ER                           & \begin{tabular}[c]{@{}c@{}}200\\ 500\\ 2000\end{tabular} & \begin{tabular}[c]{@{}l@{}}lr: 0.1\\ lr: 0.03\\ lr: 0.1\end{tabular}                                                                                 & \begin{tabular}[c]{@{}l@{}}lr: 0.1\\ lr: 0.1\\ lr: 0.1\end{tabular}                                                                                  & \begin{tabular}[c]{@{}l@{}}lr: 0.03\\ lr: 0.1\\ lr: 0.03\end{tabular}                                                                                \\ \midrule
GEM                          & \begin{tabular}[c]{@{}c@{}}200\\ 500\\ 2000\end{tabular} & \begin{tabular}[c]{@{}l@{}}lr: 0.01  $\gamma$: 1.0\\ lr: 0.01  $\gamma$: 0.5\\ lr: 0.1    $\gamma$: 0.5\end{tabular}                                                      & \begin{tabular}[c]{@{}l@{}}lr: 0.03  $\gamma$: 0.5\\ lr: 0.1    $\gamma$: 0.5\\ lr: 0.03  $\gamma$: 1.0\end{tabular}                                                      & -                                                                                                                                                    \\ \midrule
GSS                          & \begin{tabular}[c]{@{}c@{}}200\\ 500\\ 2000\end{tabular} & \begin{tabular}[c]{@{}l@{}}lr: 0.03  gmbs: 32  nb: 1\\ lr: 0.03  gmbs: 32  nb: 1\\ lr: 0.03  gmbs: 32  nb: 1\end{tabular}                            & \begin{tabular}[c]{@{}l@{}}lr: 0.03  gmbs: 32  nb: 1\\ lr: 0.03  gmbs: 32  nb: 1\\ lr: 0.03  gmbs: 32  nb: 1\end{tabular}                            & \begin{tabular}[c]{@{}l@{}}lr: 0.03  gmbs: 32  nb: 1\\ lr: 0.03  gmbs: 32  nb: 1\\ lr: 0.03  gmbs: 32  nb: 1\end{tabular}                            \\ \midrule
iCARL                        & \begin{tabular}[c]{@{}c@{}}200\\ 500\\ 2000\end{tabular} & \begin{tabular}[c]{@{}l@{}}lr: 0.1    wd: 5e-5\\ lr: 0.01  wd: 1e-5\\ lr: 0.1    wd: 1e-5\end{tabular}                                               & \begin{tabular}[c]{@{}l@{}}lr: 0.1  wd: 5e-5\\ lr: 0.1  wd: 5e-5\\ lr: 0.1  wd: 1e-5\end{tabular}                                                    & \begin{tabular}[c]{@{}l@{}}lr: 0.1    wd: 1e-5\\ lr: 0.03  wd: 1e-5\\ lr: 0.03  wd: 1e-5\end{tabular}                                                \\ \midrule
DER                          & \begin{tabular}[c]{@{}c@{}}200\\ 500\\ 2000\end{tabular} & \begin{tabular}[c]{@{}l@{}}lr: 0.03  $\alpha$: 0.2  $\beta$: 1.0 \\ lr: 0.03  $\alpha$: 0.1  $\beta$: 1.0 \\ lr: 0.03  $\alpha$: 0.2  $\beta$: 1.0\end{tabular}                             & \begin{tabular}[c]{@{}l@{}}lr: 0.03  $\alpha$: 0.5  $\beta$: 0.1 \\ lr: 0.03  $\alpha$: 0.5  $\beta$: 0.1\\ lr: 0.03  $\alpha$: 0.2  $\beta$: 0.1\end{tabular}                              & \begin{tabular}[c]{@{}l@{}}lr: 0.03  $\alpha$: 0.2  $\beta$: 0.1 \\ lr: 0.03  $\alpha$: 0.2  $\beta$: 0.1\\ lr: 0.03  $\alpha$: 0.1  $\beta$: 0.5\end{tabular}                              \\ \midrule
GCR                          & \begin{tabular}[c]{@{}c@{}}200\\ 500\\ 2000\end{tabular} & \begin{tabular}[c]{@{}l@{}}lr: 0.03  $\alpha$: 0.5  $\beta$: 0.5  $\gamma$: 0.01 \\ lr: 0.03  $\alpha$: 0.1  $\beta$: 0.1  $\gamma$: 0.1\\ lr: 0.03  $\alpha$: 0.1  $\beta$: 1.0  $\gamma$: 0.1\end{tabular}     & \begin{tabular}[c]{@{}l@{}}lr: 0.03  $\alpha$: 0.2  $\beta$: 0.1  $\gamma$: 0.01 \\ lr: 0.03  $\alpha$: 0.1  $\beta$: 0.1  $\gamma$: 0.01\\ lr: 0.03  $\alpha$: 0.2  $\beta$: 0.1  $\gamma$: 0.1\end{tabular}    & \begin{tabular}[c]{@{}l@{}}lr: 0.03  $\alpha$: 0.5  $\beta$: 0.5  $\gamma$: 0.01 \\ lr: 0.03  $\alpha$: 0.5  $\beta$: 0.1  $\gamma$: 0.01\\ lr: 0.03  $\alpha$: 0.2  $\beta$: 1.0  $\gamma$: 0.01\end{tabular}   \\ \hline
\multicolumn{5}{|c|}{\textbf{Offline Task-IL}}    
\\
\multicolumn{1}{|c|}{Method} & Buffer Size                                              & \multicolumn{1}{c|}{S-Cifar10}                                                                                                                       & \multicolumn{1}{c|}{S-Cifar100}                                                                                                                      & \multicolumn{1}{c|}{S-Tinyimg}                                                                                                                       \\ \hline
ER                           & \begin{tabular}[c]{@{}c@{}}200\\ 500\\ 2000\end{tabular} & \begin{tabular}[c]{@{}l@{}}lr: 0.01\\ lr: 0.1\\ lr: 0.03\end{tabular}                                                                                & \begin{tabular}[c]{@{}l@{}}lr: 0.03\\ lr: 0.1\\ lr: 0.1\end{tabular}                                                                                 & \begin{tabular}[c]{@{}l@{}}lr: 0.1\\ lr: 0.1\\ lr: 0.03\end{tabular}                                                                                 \\ \midrule
GEM                          & \begin{tabular}[c]{@{}c@{}}200\\ 500\\ 2000\end{tabular} & \begin{tabular}[c]{@{}l@{}}lr: 0.01  $\gamma$: 1.0\\ lr: 0.03  $\gamma$: 0.5\\ lr: 0.03  $\gamma$: 0.5\end{tabular}                                                       & \begin{tabular}[c]{@{}l@{}}lr: 0.1      $\gamma$: 0.5\\ lr: 0.03    $\gamma$: 0.5\\ lr: 0.1      $\gamma$: 0.5\end{tabular}                                               & -                                                                                                                                                    \\ \midrule
GSS                          & \begin{tabular}[c]{@{}c@{}}200\\ 500\\ 2000\end{tabular} & \begin{tabular}[c]{@{}l@{}}lr: 0.03  gmbs: 32  nb: 1\\ lr: 0.03  gmbs: 32  nb: 1\\ lr: 0.03  gmbs: 32  nb: 1\end{tabular}                            & \begin{tabular}[c]{@{}l@{}}lr: 0.03  gmbs: 32  nb: 1\\ lr: 0.03  gmbs: 32  nb: 1\\ lr: 0.03  gmbs: 32  nb: 1\end{tabular}                            & \begin{tabular}[c]{@{}l@{}}lr: 0.03  gmbs: 32  nb: 1\\ lr: 0.03  gmbs: 32  nb: 1\\ lr: 0.03  gmbs: 32  nb: 1\end{tabular}                            \\ \midrule
iCARL                        & \begin{tabular}[c]{@{}c@{}}200\\ 500\\ 2000\end{tabular} & \begin{tabular}[c]{@{}l@{}}lr: 0.1    wd: 5e-5\\ lr: 0.01  wd: 5e-5\\ lr: 0.01  wd: 5e-5\end{tabular}                                                & \begin{tabular}[c]{@{}l@{}}lr: 0.1  wd: 5e-5\\ lr: 0.1  wd: 5e-5\\ lr: 0.1  wd: 1e-5\end{tabular}                                                    & \begin{tabular}[c]{@{}l@{}}lr: 0.1    wd: 1e-5\\ lr: 0.03  wd: 1e-5\\ lr: 0.03  wd: 1e-5\end{tabular}                                                \\ \midrule
DER                          & \begin{tabular}[c]{@{}c@{}}200\\ 500\\ 2000\end{tabular} & \begin{tabular}[c]{@{}l@{}}lr: 0.03  $\alpha$: 0.2  $\beta$: 0.1 \\ lr: 0.03  $\alpha$: 0.2  $\beta$: 0.5 \\ lr: 0.03  $\alpha$: 0.2  $\beta$: 1.0\end{tabular}                             & \begin{tabular}[c]{@{}l@{}}lr: 0.03  $\alpha$: 0.1  $\beta$: 0.1 \\ lr: 0.03  $\alpha$: 0.1  $\beta$: 0.1\\ lr: 0.03  $\alpha$: 0.1  $\beta$: 0.5\end{tabular}                              & \begin{tabular}[c]{@{}l@{}}lr: 0.03  $\alpha$: 0.1  $\beta$: 0.1 \\ lr: 0.03  $\alpha$: 0.1  $\beta$: 0.5\\ lr: 0.03  $\alpha$: 0.1  $\beta$: 0.1\end{tabular}                              \\ \midrule
GCR                          & \begin{tabular}[c]{@{}c@{}}200\\ 500\\ 2000\end{tabular} & \begin{tabular}[c]{@{}l@{}}lr: 0.03  $\alpha$: 0.1  $\beta$: 0.1  $\gamma$: 0.1 \\ lr: 0.03  $\alpha$: 0.2  $\beta$: 0.5  $\gamma$: 0.01\\ lr: 0.03  $\alpha$: 0.2  $\beta$: 0.5  $\gamma$: 0.05\end{tabular}    & \begin{tabular}[c]{@{}l@{}}lr: 0.03  $\alpha$: 0.1  $\beta$: 0.1  $\gamma$: 0.01 \\ lr: 0.03  $\alpha$: 0.1  $\beta$: 0.1  $\gamma$: 0.05\\ lr: 0.03  $\alpha$: 0.1  $\beta$: 0.1  $\gamma$: 0.1\end{tabular}    & \begin{tabular}[c]{@{}l@{}}lr: 0.03  $\alpha$: 0.1  $\beta$: 0.5  $\gamma$: 0.01 \\ lr: 0.03  $\alpha$: 0.1  $\beta$: 1.0  $\gamma$: 0.01\\ lr: 0.03  $\alpha$: 0.1  $\beta$: 1.0  $\gamma$: 0.01\end{tabular}   \\ \hline
\multicolumn{5}{|c|}{\textbf{Online Streaming}}                                                                                                                                                                                                                                                                                                                                                                                                                                                                                                              \\
\multicolumn{1}{|c|}{Method} & Buffer Size                                              & \multicolumn{1}{c|}{S-Cifar10}                                                                                                                       & \multicolumn{1}{c|}{S-Cifar100}                                                                                                                      & \multicolumn{1}{c|}{S-Tinyimg}                                                                                                                       \\ \hline
ER                           & \begin{tabular}[c]{@{}c@{}}200\\ 500\\ 2000\end{tabular} & \begin{tabular}[c]{@{}l@{}}lr: 0.03\\ lr: 0.01\\ lr: 0.01\end{tabular}                                                                               & \begin{tabular}[c]{@{}l@{}}lr: 0.01\\ lr: 0.01\\ lr: 0.03\end{tabular}                                                                               & \begin{tabular}[c]{@{}l@{}}lr: 0.1\\ lr: 0.01\\ lr: 0.01\end{tabular}                                                                                \\ \midrule
GEM                          & \begin{tabular}[c]{@{}c@{}}200\\ 500\\ 2000\end{tabular} & \begin{tabular}[c]{@{}l@{}}lr: 0.1    $\gamma$: 0.5\\ lr: 0.03  $\gamma$: 1.0\\ lr: 0.1    $\gamma$: 1.0\end{tabular}                                                     & \begin{tabular}[c]{@{}l@{}}lr: 0.03  $\gamma$: 0.5\\ lr: 0.1    $\gamma$: 0.5\\ lr: 0.03  $\gamma$: 1.0\end{tabular}                                                      & \begin{tabular}[c]{@{}l@{}}lr: 0.03  $\gamma$: 0.5\\ lr: 0.03  $\gamma$: 1.0\\ lr: 0.03  $\gamma$: 1.0\end{tabular}                                                       \\ \midrule
GSS                          & \begin{tabular}[c]{@{}c@{}}200\\ 500\\ 2000\end{tabular} & \begin{tabular}[c]{@{}l@{}}lr: 0.03  gmbs: 32  nb: 1\\ lr: 0.03  gmbs: 32  nb: 1\\ lr: 0.03  gmbs: 32  nb: 1\end{tabular}                            & \begin{tabular}[c]{@{}l@{}}lr: 0.03  gmbs: 32  nb: 1\\ lr: 0.03  gmbs: 32  nb: 1\\ lr: 0.03  gmbs: 32  nb: 1\end{tabular}                            & \begin{tabular}[c]{@{}l@{}}lr: 0.1  gmbs: 32  nb: 1\\ lr: 0.1  gmbs: 32  nb: 1\\ lr: 0.1  gmbs: 32  nb: 1\end{tabular}                               \\ \midrule
iCARL                        & \begin{tabular}[c]{@{}c@{}}200\\ 500\\ 2000\end{tabular} & \begin{tabular}[c]{@{}l@{}}lr: 0.1  wd: 1e-5\\ lr: 0.1  wd: 1e-5\\ lr: 0.1  wd: 1e-5\end{tabular}                                                    & \begin{tabular}[c]{@{}l@{}}lr: 0.1  wd: 5e-5\\ lr: 0.1  wd: 5e-5\\ lr: 0.1  wd: 5e-5\end{tabular}                                                    & \begin{tabular}[c]{@{}l@{}}lr: 0.1  wd: 5e-5\\ lr: 0.1  wd: 5e-5\\ lr: 0.1  wd: 5e-5\end{tabular}                                                    \\ \midrule
MIR                          & \begin{tabular}[c]{@{}c@{}}200\\ 500\\ 2000\end{tabular} & \begin{tabular}[c]{@{}l@{}}lr: 0.03\\ lr: 0.03\\ lr: 0.03\end{tabular}                                                                               & \begin{tabular}[c]{@{}l@{}}lr: 0.03\\ lr: 0.03\\ lr: 0.03\end{tabular}                                                                               & \begin{tabular}[c]{@{}l@{}}lr: 0.03\\ lr: 0.03\\ lr: 0.03\end{tabular}                                                                               \\ \midrule
DER                          & \begin{tabular}[c]{@{}c@{}}200\\ 500\\ 2000\end{tabular} & \begin{tabular}[c]{@{}l@{}}lr: 0.03  $\alpha$: 0.1  $\beta$: 1.0 \\ lr: 0.01  $\alpha$: 0.2  $\beta$: 2.0 \\ lr: 0.01  $\alpha$: 0.2  $\beta$: 1.0\end{tabular}                             & \begin{tabular}[c]{@{}l@{}}lr: 0.03  $\alpha$: 0.5  $\beta$: 0.5 \\ lr: 0.03  $\alpha$: 1.0  $\beta$: 0.5\\ lr: 0.01  $\alpha$: 1.0  $\beta$: 2.0\end{tabular}                              & \begin{tabular}[c]{@{}l@{}}lr: 0.03    $\alpha$: 0.2  $\beta$: 2.0 \\ lr: 0.005  $\alpha$: 1.0  $\beta$: 3.0\\ lr: 0.01    $\alpha$: 1.0  $\beta$: 3.5\end{tabular}                         \\ \midrule
GCR                          & \begin{tabular}[c]{@{}c@{}}200\\ 500\\ 2000\end{tabular} & \begin{tabular}[c]{@{}l@{}}lr: 0.03    $\alpha$: 0.2  $\beta$: 1.0  $\gamma$: 1.0 \\ lr: 0.005  $\alpha$: 0.5  $\beta$: 3.5  $\gamma$: 1.0\\ lr: 0.03    $\alpha$: 0.1  $\beta$: 0.5  $\gamma$: 1.5\end{tabular} & \begin{tabular}[c]{@{}l@{}}lr: 0.005  $\alpha$: 1.0  $\beta$: 3.5  $\gamma$: 1.0 \\ lr: 0.01    $\alpha$: 0.2  $\beta$: 3.0  $\gamma$: 1.5\\ lr: 0.01    $\alpha$: 1.0  $\beta$: 2.0  $\gamma$: 1.0\end{tabular} & \begin{tabular}[c]{@{}l@{}}lr: 0.01    $\alpha$: 1.0  $\beta$: 3.0  $\gamma$: 0.1 \\ lr: 0.005  $\alpha$: 1.0  $\beta$: 3.5  $\gamma$: 1.0\\ lr: 0.01    $\alpha$: 1.0  $\beta$: 3.0  $\gamma$: 1.5\end{tabular} \\ \hline
\end{tabular}}
\caption{Hyperparameter values obtained from the grid search.}
\label{tab:best_values_table}
\end{table*}
Table \ref{tab:best_values_table} shows the hyperparameter values selected from the grid search that were used in our experiments.

\subsection{Hyperparameter Search Space}
\begin{table*}[ht]
\centering
\begin{tabular}{|l|l|l|l|}
\hline
\multicolumn{1}{|c|}{Method} & \multicolumn{1}{c|}{Parameters}                                            & \multicolumn{1}{c|}{Offline}                                                                                        & \multicolumn{1}{c|}{Online}                                                                                                                  \\ \hline
ER                           & lr                                                                         & [0.01, 0.03, 0.1]                                                                                                   & [0.01, 0.03, 0.1]                                                                                                                            \\ \midrule
GEM                          & \begin{tabular}[c]{@{}l@{}}lr\\ $\gamma$\end{tabular}                      & \begin{tabular}[c]{@{}l@{}}$[0.01, 0.03, 0.1]$\\ $[0.5, 1.0]$\end{tabular}                                              & \begin{tabular}[c]{@{}l@{}}$[0.01, 0.03, 0.1]$\\ $[0.5, 1.0]$\end{tabular}                                                                       \\ \midrule
GSS                          & lr                                                                         & $[0.01, 0.03]$                                                                                                        & $[0.005, 0.01, 0.03, 0.1]$                                                                                                                     \\ \midrule
iCARL                        & \begin{tabular}[c]{@{}l@{}}lr\\ wd\end{tabular}                            & \begin{tabular}[c]{@{}l@{}}$[0.01, 0.03, 0.1]$\\ $[1e-5, 5e-5]$\end{tabular}                                            & \begin{tabular}[c]{@{}l@{}}$[0.01, 0.03, 0.1]$\\ $[1e-5, 5e-5]$\end{tabular}                                                                     \\ \midrule
MIR                          & lr                                                                         & -                                                                                                                   & $[0.01, 0.03]$                                                                                                                                 \\ \midrule
DER                          & \begin{tabular}[c]{@{}l@{}}lr\\ $\alpha$\\ $\beta$\end{tabular}            & \begin{tabular}[c]{@{}l@{}}[0.03]\\ $[0.1, 0.2, 0.5, 1.0]$\\ $[0.1, 0.5, 1.0]$\end{tabular}                             & \begin{tabular}[c]{@{}l@{}}$[0.005,0.01, 0.03]$\\ $[0.1, 0.2, 0.5, 1.0]$\\ $[0.1, 0.5, 1.0, 2.0, 3.0, 3.5]$\end{tabular}                           \\ \midrule
GCR                          & \begin{tabular}[c]{@{}l@{}}lr\\ $\alpha$\\ $\beta$\\ $\gamma$\end{tabular} & \begin{tabular}[c]{@{}l@{}}$[0.03]$\\ $[0.1, 0.2, 0.5, 1.0]$\\ $[0.1, 0.5, 1.0]$\\ $[0, 0.01, 0.05, 0.1, 0.2]$\end{tabular} & \begin{tabular}[c]{@{}l@{}}$[0.005,0.01, 0.03]$\\ $[0.1, 0.2, 0.5, 1.0]$\\ $[0.1, 0.5, 1.0, 2.0, 3.0, 3.5]$\\ $[0, 0.1, 0.5, 1.0, 1.5]$\end{tabular} \\ \hline
\end{tabular}
\caption{Hyperparameter Search Space.}
\label{tab:grid_search_table}
\end{table*}
Table \ref{tab:grid_search_table} shows the hyperparameter search space for offline and online setting on which grid search was done.
\end{document}